%% file: main.tex
\documentclass{article}

\usepackage{microtype}
\usepackage{graphicx}
\usepackage{subfigure}
\usepackage{booktabs} 
\usepackage{makecell}
\usepackage{lipsum}
\setlipsum{%
  par-before = \begingroup\color{lightgray},
  par-after = \endgroup
}
\usepackage{listings}
\lstdefinestyle{Python}{
  language=Python,
  basicstyle=\ttfamily,
  keywordstyle=\color{cyan},
  commentstyle=\color{gray},
  stringstyle=\color{purple},
  showstringspaces=false,
  showspaces=false,
  breaklines=true,
  breakatwhitespace=true,
  tabsize=4
}
\newcommand{\code}{\texttt}

\usepackage{hyperref}

\usepackage[accepted]{icml_codeml_2025}

\usepackage{amsmath}
\usepackage{amssymb}
\usepackage{mathtools}
\usepackage{amsthm}

\usepackage[capitalize,noabbrev]{cleveref}

\theoremstyle{plain}

\theoremstyle{definition}

\theoremstyle{remark}

\usepackage[textsize=tiny]{todonotes}

\icmltitlerunning{TorchAO: PyTorch-native Training-to-Serving Model Optimization}

\begin{document}

\twocolumn[
\icmltitle{TorchAO: PyTorch-Native Training-to-Serving Model Optimization}



\icmlsetsymbol{equal}{*}

\begin{icmlauthorlist}
\icmlauthor{Andrew Or}{meta}
\icmlauthor{Apurva Jain}{meta}
\icmlauthor{Daniel Vega-Myhre}{meta}
\icmlauthor{Jesse Cai}{meta}
\icmlauthor{Charles David Hernandez}{meta}
\icmlauthor{Zhenrui Zhang}{meta}
\icmlauthor{Driss Guessous}{meta}
\icmlauthor{Vasiliy Kuznetsov}{meta}
\icmlauthor{Christian Puhrsch}{meta}
\icmlauthor{Mark Saroufim}{meta}
\icmlauthor{Supriya Rao}{meta}
\icmlauthor{Thien Tran}{independent}
\icmlauthor{Aleksandar Samardžić}{openteams}

\end{icmlauthorlist}

\icmlaffiliation{meta}{Meta Platforms Inc.}
\icmlaffiliation{independent}{Independent}
\icmlaffiliation{openteams}{OpenTeams Inc.}

\icmlcorrespondingauthor{Andrew Or}{andrewor@meta.com}
\icmlcorrespondingauthor{Supriya Rao}{supriyar@meta.com}

\icmlkeywords{Machine Learning, ICML}

\vskip 0.3in
]



\printAffiliationsAndNotice{}  

\begin{abstract}
We present TorchAO, a PyTorch-native model optimization framework leveraging
quantization and sparsity to provide an end-to-end, training-to-serving workflow
for AI models. TorchAO supports a variety of popular model optimization techniques,
including FP8 quantized training, quantization-aware training (QAT), post-training
quantization (PTQ), and 2:4 sparsity, and leverages a novel tensor subclass
abstraction to represent a variety of widely-used, backend agnostic low precision
data types, including INT4, INT8, FP8, MXFP4, MXFP6, and MXFP8.

TorchAO integrates closely with the broader ecosystem at each step of the model
optimization pipeline, from pre-training (TorchTitan~\cite{torchtitan}) to fine-tuning
(TorchTune~\cite{torchtune}, Axolotl~\cite{axolotl}) to serving (HuggingFace~\cite{hf_transformers}, vLLM~\cite{vllm}, SGLang~\cite{sglang},
ExecuTorch~\cite{executorch}), connecting an otherwise fragmented space in a single,
unified workflow. TorchAO has enabled recent launches of the quantized Llama 3.2
1B/3B~\cite{quantizedllama3.2} and LlamaGuard3-8B models~\cite{llamaguard}
and is open-source at \url{https://github.com/pytorch/ao/}.
\end{abstract}

\input{intro}
\input{e2e_flow1}
\input{e2e_flow2}
\input{eval}
\input{conclusion}

\section*{Acknowledgement}
We express our gratitude towards our many open-source contributors and
collaborators, including but certainly not limited to
Hicham Badri, Sayak Paul, Aryan V S, Marc Sun, Matthew Douglas, Salman Mohammadi,
Lianmin Zheng, Michael Goin, Daniel Han, Peter Yeh, Jeff Daily,
Pawan Jayakumar, Jerome Ku, Tobias van der Werff, Vaishnavi Gupta, Andreas Köpf, Diogo Venâncio,
Leslie Fang, Weiwen Xia, Yi Liu, Xuan Liao, Yanbing Jiang, Xiao Wang, Sanchit Jain, Hengyu Meng, Devang Choudhary,
Ethan Petersen, Martin Cala, Chip Smith,
Scott Roy, Digant Desai, Kimish Patel, Manual Candales, Mengwei Liu, Songhao Jia, Chen Lai, Zeyu Song, Yanan Cao,
Andrey Talman, Huy Do, Catherine Lee, Svetlana Karslioglu, Evan Smothers,
Will Feng, Will Constable, Ke Wen, Tianyu Liu, Gokul Nadathur, Tristan Rice, Shuqi Yang
Lisa Jin, Zechun Liu, Tijmen Blankevoort, Hanxian Huang, Luca Wehrstedt,
Daniel Haziza, Timothy Chou, Dhruv Choudhary, Francisco Massa, Jiecao Yu,
Geonhwa Jeong, Patrick Labatut, Josh Fromm, Less Wright, and Hamid Shojanazeri.
We also thank Joe Isaacson, Jane Xu, and Scott Roy for your early feedback on the paper.


\bibliography{main}
\bibliographystyle{icml2025}

\input{appendix}
\end{document}

%% file: intro.tex
\section{Introduction}

Large Language Models (LLMs) have been at the forefront of content
creation, text summarization, chatbots, and code generation, among
a wide variety of other use cases. However, such capabilities often
require substantial infrastructure, as seen in top-performing
models such as Qwen3 (235B parameters)~\cite{qwen3}, DeepSeek-v3 (671B)~\cite{deepseek},
Llama 3.1 (405B)~\cite{llama3.1}, and Llama 4 Behemoth (2T)~\cite{llama4}.

The computational costs and memory footprint of these models
pose significant challenges in every step of the LLM pipeline, from
training to fine-tuning to serving. For instance, training Llama 3.1
took 30.84M GPU hours on 16K H100 GPUs \cite{llama3.1}, and even
serving the model in its original BF16 precision requires at least
800GB aggregate memory just to fit the model, exceeding the memory
limitations of a single server with 8 H100 GPUs. Even at the smaller
1-8B parameter scale, reducing the sizes of these models is important
for deploying them in resource-constrained environments such as
mobile and edge devices.

However, the existing LLM optimization pipeline is highly fragmented.
For instance, a researcher may pre-train their model using mixed FP8/BF16
precision support in Transformer Engine~\cite{te_fp8}, load the pre-trained
model into Unsloth~\cite{unsloth} or Axolotl~\cite{axolotl} for further
fine-tuning, perform quantization using bitsandbytes~\cite{llm_int8}
before finally serving the model using llama.cpp~\cite{llama.cpp}.
In each step, the user may need to manually convert the model format
(e.g. from HuggingFace's safetensors to GGUF in llama.cpp), and the
quantization schemes may diverge from the ones used in previous steps
with subtle discrepancies.

\begin{figure}[t!]
  \centering
  \advance\leftskip-0.2cm
  \includegraphics[scale=0.17]{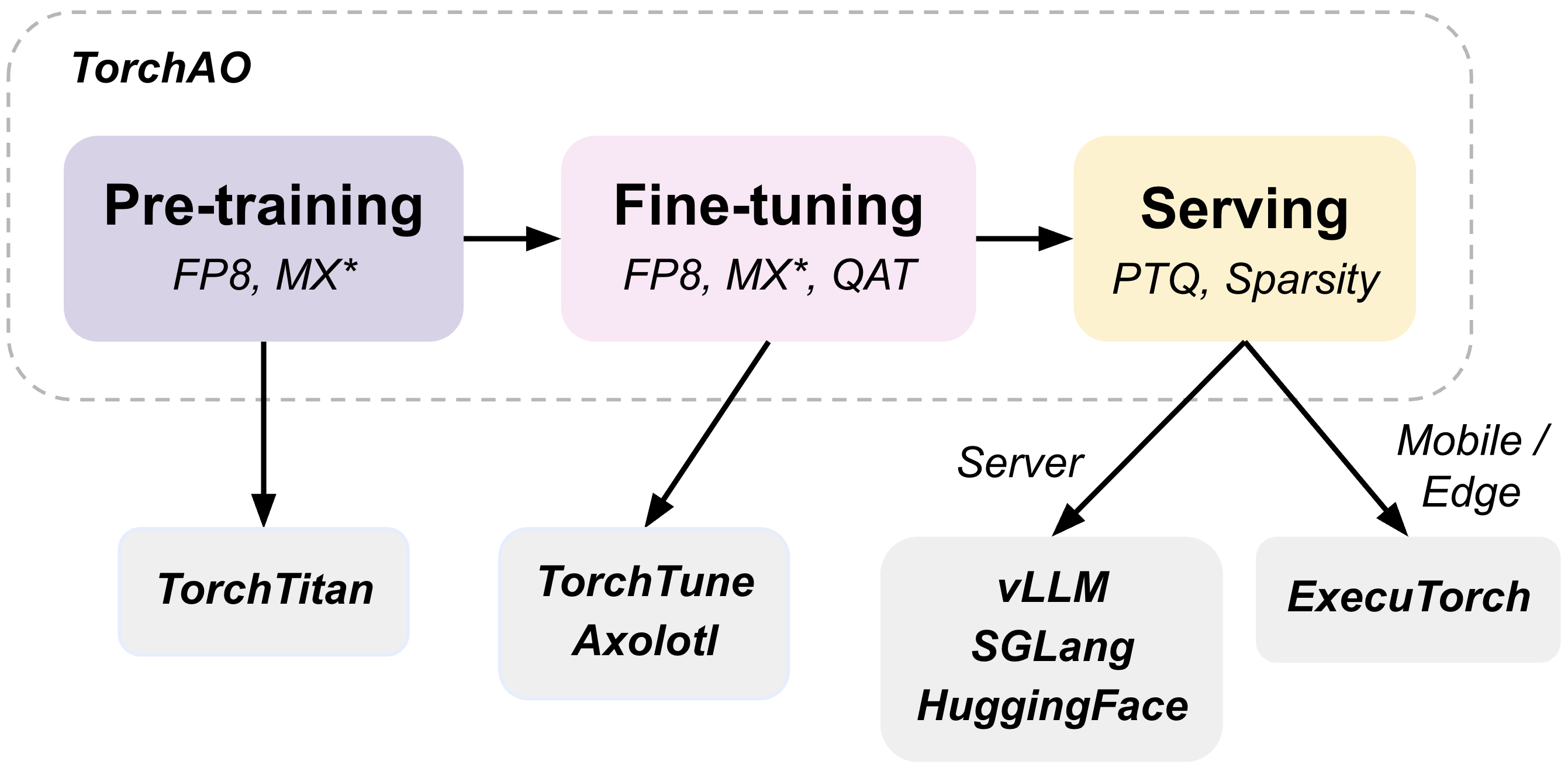}
  \vspace{-2mm}
  \caption{\textbf{TorchAO optimization workflow.} TorchAO is closely
    integrated with popular pre-training, fine-tuning, serving, and
    model-definition frameworks to provide a seamless, PyTorch-native
    end-to-end workflow for users to optimize their models.
    $^\ast$MX training is a prototype feature.}
  \label{image:torchao_overview}
  \vspace{-3mm}
\end{figure}

\begin{figure}
  \centering
  \advance\leftskip-0.2cm
  \includegraphics[scale=0.25]{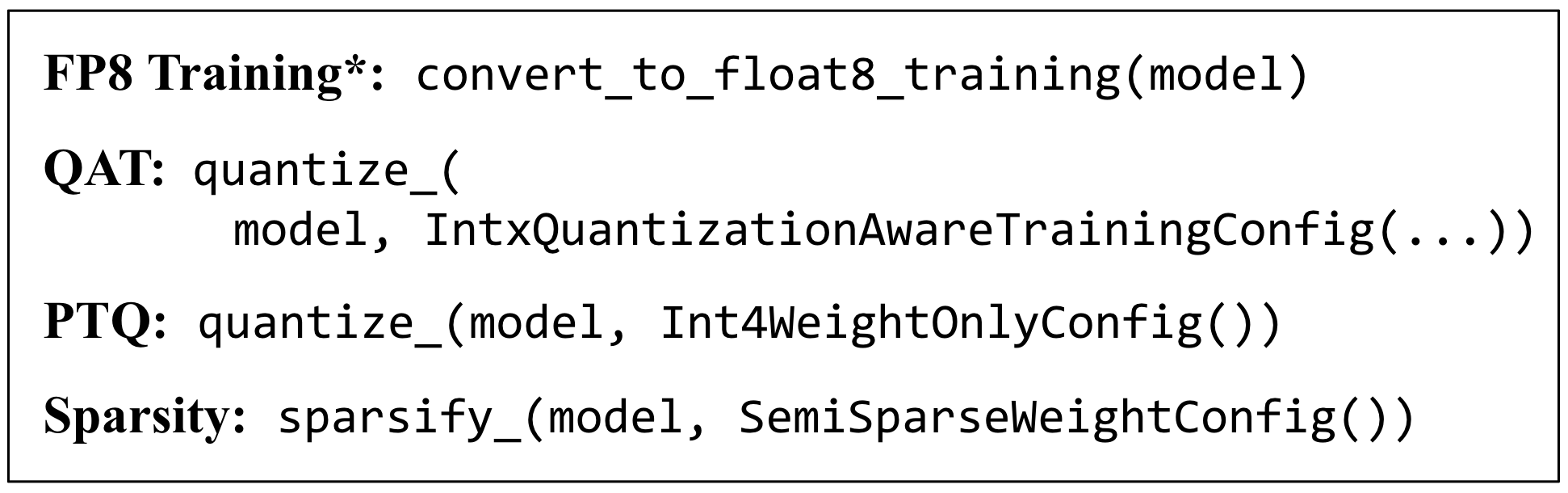}
  \vspace{-6mm}
  \caption{\textbf{TorchAO APIs.}
    Users can optimize their models using the above one-line
    transformations. $^\ast$API subject to change.
  }
  \label{image:torchao_apis}
  \vspace{-3mm}
\end{figure}

We present TorchAO, a PyTorch-native model optimization
framework leveraging quantization and sparsity to provide an end-to-end,
training-to-serving workflow for AI models. TorchAO integrates closely with
the broader ecosystem at each step of the model optimization pipeline
(Figure~\ref{image:torchao_overview}):

\begin{itemize}
\vspace{-2mm}
\item\textbf{Pre-training.} TorchAO's FP8 training support
(\S\ref{sec:fp8_training}) composes natively with PyTorch features
such as \code{torch.compile}, autograd, FSDP2, and tensor parallelism
support, reaping throughput gains ($\sim$1.5x at 405B scale)~\cite{pytorch_float8_fsdp2_blog}
with virtually no change in model quality when integrated into
TorchTitan's pre-training~\cite{torchtitan}.
\item\textbf{Fine-tuning.} Quantization-Aware Training (QAT) is a
popular technique for mitigating quantization degradation by simulating
quantization numerics during training. TorchAO's QAT support can recover
up to 96\% of the degradation in quantized accuracy~\cite{torchao_qat_blog}
and can be composed with LoRA~\cite{lora} to improve the training
throughput by 1.89x compared to vanilla QAT~\cite{qat_lora, quantizedllama3.2}.
Additionally, TorchAO also provides the NF4 data type for QLoRA~\cite{qlora}
to further reduce resource requirements during training.

TorchTune~\cite{torchtune} natively integrates TorchAO's QAT, NF4,
and FP8 training support into its fine-tuning recipes. Axolotl~\cite{axolotl}
also provides fine-tuning workflows that leverage TorchAO QAT,
including one that composes QAT with Direct Preference Optimization (DPO)~\cite{dpo}.
\vspace{-4mm}
\item\textbf{Serving.} TorchAO provides flexible support for
a wide variety of backend agnostic Post-Training Quantization (PTQ)
schemes for model optimization before serving.
For server backends, TorchAO's PTQ support is integrated
as an optional quantization transformation before serving in
SGLang~\cite{torchao_sglang} or vLLM~\cite{torchao_vllm}. For edge
and mobile backends, users can lower their TorchAO quantized models
in ExecuTorch~\cite{torchao_executorch, quantizedllama3.2}, which
provides lightweight runtimes with static memory planning to reduce
performance and power overheads, and serve their models on-device
using our custom quantized ARM CPU and metal kernels~\cite{torchao_kernels}.
\end{itemize}

TorchAO is also closely integrated with HuggingFace Transformers~\cite{hf_transformers}
and Diffusers~\cite{hf_diffusers}, two popular model-definition
frameworks for state-of-the-art machine learning models~(Listing~\ref{lst:torchao_hf}).
More specifically, TorchAO is natively integrated as one of the
quantization backends in HuggingFace, enabling users to optionally
apply post-training quantization when loading their models from
the HuggingFace Hub. Our integration also supports serialization
through native HuggingFace APIs such as \texttt{save\_pretrained},
\texttt{load\_pretrained}, and \texttt{push\_to\_hub}, further
enabling users to seamlessly save and load their quantized models for
future inference or generation on their desired serving frameworks.

\begin{lstlisting}[
    style=Python,
    float=t!,
    basicstyle=\ttfamily\small,
    frame=single,
    captionpos=b,
    caption={
      \textbf{TorchAO as a quantization backend in HuggingFace.}
      TorchAO quantization can be applied to any model on HuggingFace Hub
      through the $\texttt{TorchAoConfig}$. Users can then serialize and
      deserialize their quantized models and upload them to HuggingFace Hub
      at ease with native HuggingFace APIs: $\texttt{save\_pretrained}$,
      $\texttt{load\_pretrained}$, and $\texttt{push\_to\_hub}$.
    },
    label={lst:torchao_hf},
    belowskip=-5pt,
]
from torchao.quantization import (
  Int4WeightOnlyConfig,
)
from transformers import (
  AutoModelForCausalLM,
  TorchAoConfig,
)
config = Int4WeightOnlyConfig()
config = TorchAoConfig(quant_type=config)
mod = AutoModelForCausalLM.from_pretrained(
  "meta-llama/Llama-3.2-3B",
  device_map="auto",
  torch_dtype=torch.bfloat16,
  quantization_config=config,
)
mod.push_to_hub(
  f"{user_id}/Llama-3.2-3B-int4",
  safe_serialization=False,
)
\end{lstlisting}

Alternatively, users can simply call TorchAO's one-line APIs directly to
optimize their models~(Figure~\ref{image:torchao_apis}). For instance,
users may wish to use a custom training loop or a training framework
not yet integrated with TorchAO for FP8 pre-training or QAT fine-tuning.
For more detailed usage, please refer to Appendix~\ref{sec:appendix_apis}.

In the ensuing sections, we will walk through two end-to-end example
workflows that leverage different model optimization techniques in
TorchAO targeting different backends. In the first workflow, we will
use FP8 training and inference as our unifying theme and target server
GPUs as the serving backend~(\S\ref{sec:e2e_flow1}). In the second workflow,
we will leverage QAT to mitigate quantization degradation and lower our
model to the XNNPACK~\cite{xnnpack} backend, which provides optimized
ARM kernels used by the most popular Android and iOS phones~(\S\ref{sec:e2e_flow2}).

%% file: e2e_flow1.tex
\section{Workflow: FP8 Targeting Server GPUs}
\label{sec:e2e_flow1}

The first end-to-end workflow leverages TorchAO's FP8 training
support to pre-train and fine-tune the model using TorchTitan
and TorchTune respectively, and then serves the trained model
in vLLM using the same FP8 configurations.

Typically, pre-training is a time consuming step that trains
on large, general datasets like C4~\cite{c4}, while fine-tuning
is a separate step that adapts the pre-trained model to more
domain specific tasks. In this example workflow, both training
steps leverage TorchAO's dynamic FP8 training with tensorwise
scaling to speed up the training process. In each
step of the workflow, users may optionally upload their pre-trained
or fine-tuned checkpoints to HuggingFace hub for others in the
community to continue fine-tuning the models or directly use them
for inference~(Listing~\ref{lst:fp8_flow}).

\begin{lstlisting}[
    style=Python,
    float=t!,
    basicstyle=\ttfamily\small,
    frame=single,
    captionpos=b,
    caption={\textbf{Example end-to-end FP8 flow.} The user leverages
    TorchAO's FP8 training support to pre-train and fine-tune their model
    using TorchTitan and TorchTune respectively, then (optionally) uploads
    their trained model to HuggingFace hub and serves it using vLLM.
    Detailed FP8 training options are omitted for brevity.},
    label={lst:fp8_flow},
    belowskip=-8pt,
]
# Pre-training in FP8 with TorchTitan
torchtitan/run_train.sh
  --training.compile
  --model.converters="float8"

# Fine-tuning in FP8 with TorchTune
tune run --nnodes 1 --nproc_per_node 8
  full_finetune_distributed
  --config llama3.1/8B_full
    enable_fp8_training="true"

# Upload FP8 model to HuggingFace hub
# (Optional, not shown)

# Serve FP8 model through vLLM
vllm serve
  <HF_USER_ID>/Llama-3.1-8B-Instruct-fp8
  --tokenizer meta-llama/Llama-3.1-8B
\end{lstlisting}

\subsection{FP8 Training}
\label{sec:fp8_training}

TorchAO's FP8 training dynamically casts activations, weights,
and gradients to FP8 and leverages specialized GEMM kernels
to take advantage of the FP8 tensor cores~\cite{h100} in the underlying GPUs.
This technique is most useful for training models in which the
majority of GEMM operations are large enough such that the speedup
achieved by using FP8 tensor cores is greater than the overhead of
dynamic quantization~(Appendix~\ref{sec:appendix_float8_microbenchmarks}).

We support three FP8 training scaling recipes, each with different
throughput/accuracy trade-offs: tensorwise, rowwise, and rowwise with
high precision \code{grad\_weight} (see Appendix~\ref{sec:appendix_fp8_scaling}
for full recipe details). Combined with \code{torch.compile},
tensorwise scaling achieved training throughput speedups of up to 1.5x at
405B parameter scale on 512 H100 GPUs~\cite{pytorch_float8_fsdp2_blog},
and rowwise scaling achieved speedups of up to 1.43x at 70B parameter
scale on 1920 H200 GPUs~\cite{pytorch_float8_crusoe_blog},
compared to training with BF16.

TorchAO's FP8 training leverages the tensor subclass
abstraction~\cite{tensor_subclasses} to compose with PyTorch autograd
and PyTorch distributed. Our FP8 training support is integrated into
both TorchTitan and TorchTune, so all of the above scaling recipes
can be used for pre-training and fine-tuning with minimal setup~(Listing~\ref{lst:fp8_flow}).

For further throughput gains, users can leverage asynchronous tensor
parallelism~\cite{async_tp}, a compiler driven approach to overlapping
the compute and communications in tensor parallel, via using
SymmetricMemory APIs~\cite{symmetric_memory} for copy-engine based
communications instead of SM-based ones, and micro-pipelining the
computations. TorchAO's FP8 tensorwise and rowwise recipes are composable
with asynchronous tensor parallelism, yielding up to 17\% additional
training throughput improvement, depending on factors like model size and
activation checkpointing strategy~\cite{torchtitan_async_tp}.

\begin{table}[t]
    \centering
    \small
    \advance\leftskip-0.1cm
    \setlength{\tabcolsep}{2pt}
    \begin{tabular}{llll}
        \toprule
        Quantization & \thead{Output token \\ throughput (tok/s)} & \thead{Time per output \\ token (ms)} & \thead{Inter-token \\ latency (ms)} \\
        \midrule
        None (BF16) & 103.6 (+0\%) & 9.50 (+0\%) & 9.47 (+0\%) \\
        FP8 tensorwise & 132.8 (+28.2\%) & 7.48 (-21.2\%) & 7.47 (-21.1\%) \\
        \bottomrule
    \end{tabular}
    \vspace{-2mm}
    \caption{\textbf{Serving FP8 pre-trained and fine-tuned Llama3.1-8B on vLLM.}
    Serving this model in FP8 vs in the original precision (BF16) saw a 28\%
    increase in throughput and a 21\% reduction in latency. Clients in both
    experiments used the ShareGPT dataset and number of prompts = 1.}
    \vspace{-4mm}
    \label{tab:vllm_serving}
\end{table}

\subsection{Post-Training Optimization}

Similar to FP8 training, both post-training quantization (PTQ) and sparsity
support in TorchAO leverage PyTorch's tensor subclass abstraction
to provide native composability with other PyTorch features and seamless
serialization support. TorchAO's PTQ supports a wide range of popular data
types, including INT4, INT8, FP8, MXFP4, MXFP6, and MXFP8, and can be lowered
to efficient kernels across different backends such as CUDA and ARM CPU.
TorchAO's PTQ is also integrated with GemLite~\cite{gemlite} to leverage
specialized Triton kernels~\cite{triton} to further speed up inference by
1.1-2x across different batch sizes and tensor parallel sizes~\cite{torchao_sglang}.

TorchAO's sparsity support accelerates inference by leveraging hardware
support for sparse matrix multiplications offered by modern NVIDIA GPUs~\cite{mishra2021accelerating}.
Our benchmarks have shown up to 1.3x speedup with relative model accuracy
of ~91-100\% on ViT models compared to the non-sparse baseline. TorchAO
provides APIs and kernels for different sparsity techniques including
sparse marlin 2:4~\cite{marlin}, 2:4 sparsity, block sparsity, INT8
dynamic quantization + 2:4 sparsity, and rowwise FP8 + 2:4 sparsity for
weights and activations~\cite{2_4_sparsity}.

For detailed quantization and sparsity benchmarks, refer to the respective READMEs~\cite{torchao_ptq_benchmarks,torchao_sparsity_benchmarks}.

\subsection{Serving}

TorchAO's PTQ support is closely integrated with popular
model serving frameworks such as vLLM~\cite{vllm} and SGLang~\cite{sglang} as
one of the quantization backends. In particular, our FP8 inference support uses
the same configurations as FP8 training to provide consistent end-to-end numerics
across different steps in the workflow. Initial benchmarks on serving Llama3.1-8B
in FP8 vs in the original precision (BF16) demonstrates a 28\% increase in
throughput and 21\% reduction in latency~(Table~\ref{tab:vllm_serving}).

%% file: e2e_flow2.tex
\section{Workflow: QAT Targeting Mobile Devices}
\label{sec:e2e_flow2}

The second end-to-end workflow leverages TorchAO's
Quantization-Aware Training (QAT) support to fine-tune
the model using TorchTune, and then lowers the model
to the XNNPACK backend using ExecuTorch, enabling the
model to be accessed on mobile phones and wearable
devices such as smart glasses.

%
%
%

\begin{lstlisting}[
    style=Python,
    float=t!,
    basicstyle=\ttfamily\small,
    frame=single,
    captionpos=b,
    caption={\textbf{Example end-to-end QAT Targeting Mobile Devices.} The user leverages
    TorchTune's QAT and PTQ support to fine-tune and quantize Llama3.1-8B, then exports
    the model using ExecuTorch and serves it as an INT4 model on CPU.
    Detailed export flags are omitted for brevity.},
    label={lst:qat_flow},
    belowskip=-5pt,
]
# Fine-tuning using torchtune's QAT recipe
tune run --nnodes 1 --nproc_per_node 4
  qat_distributed
  --config llama3/8B_qat_full

# Lower model to executorch
python -m examples.models.llama.export_llama
  --checkpoint <checkpoint.pth>
  --params <params.json>
  --use_sdpa_with_kv_cache
  --xnnpack
  --use_kv_cache
  --embedding-quantize 4,32
  --quantization-mode 8da4w

# Serve on mobile device
# Build llama runner and run on Android
adb shell "cd /data/local/tmp/llama && ./llama_main --model_path <model.pte> --tokenizer_path <tokenizer.model> --prompt 'What is the capital of France?' --seq_len 120" --warmup=1
\end{lstlisting}

\begin{table*}[ht]
    \centering
    \small
    \setlength{\tabcolsep}{4pt}
    \begin{tabular}{lllll}
        \toprule
        Model & \thead{Quantized hellaswag \\ accuracy} & \thead{Quantized wikitext \\ word perplexity} & \thead{Training \\ throughput (tok/s)} & \thead{Training peak \\ memory (GB)} \\
        \midrule
        Llama3-8B & 47.0\% (57.1\% BF16) & 26.270 (9.422 BF16) & 480.3 (+0\%) & 17.6 (+0\%) \\
        Llama3-8B (QAT) & 52.8\% (recovered 57.8\%) & 12.312 (\textbf{recovered 82.8\%}) & 323.0 (-32.7\%) & 32.9 (+86.8\%) \\
        Llama3.1-8B & 51.8\% (57.9\% BF16) & 18.628 (9.164 BF16) & 492.4 (+0\%) & 17.7 (+0\%) \\
        Llama3.1-8B (QAT) & 55.5\% (recovered 60.0\%) & 10.901 (recovered 81.6\%) & 323.0 (-34.4\%) & 33.0 (+86.5\%) \\
        Llama3.2-3B & 46.8\% (51.7\% BF16) & 17.461 (12.051 BF16) & 1408.8 (+0\%) & 13.8 (+0\%) \\
        Llama3.2-3B (QAT) & 50.2\% (\textbf{recovered 69.8\%}) & 13.220 (recovered 78.4\%) & 737.7 (-47.6\%) & 14.5 (+5.24\%) \\
        \bottomrule
    \end{tabular}
    \caption{\textbf{Quantization-Aware Training (QAT) on Llama3 models,
    fine-tuned on OpenAssistant Conversations Dataset (OASST1)~\cite{oasst1}.}
    For this workload, QAT can recover up to 69.8\% of the degradation in
    quantized hellaswag accuracy and 82.8\% of the degradataion in quantized
    wikitext word perplexity.}
    \label{tab:qat}
\end{table*}

\begin{table}
    \centering
    \small
    \setlength{\tabcolsep}{4pt}
    \begin{tabular}{llcccc}
    \hline
    Scaling & \thead{Peak Mem \\ (GB)} & \thead{Median \\ tok/s} & Speedup \\
    \hline
    None (BF16) & 47.65 & 6150 & 1.0 \\
    tensorwise + FP8 all-gather & 47.77 & 7689.5 & 1.25 \\
    rowwise + BF16 all-gather & 47.79 & 6768 & 1.10 \\
    \hline
    \end{tabular}
    \caption{\textbf{FP8 pre-training on Llama3-8B using TorchTitan.}
    Tensorwise scaling with FP8 all-gather achieves 1.25x faster
    training throughput on this model with on par memory usage.}
    \label{tab:fp8}
\end{table}

\begin{table}[h]
    \centering
    \small
    \setlength{\tabcolsep}{4pt}
    \begin{tabular}{lcccc}
        \toprule
        \thead{Quantization \\ Technique} & Acc & \thead{Word \\ perplexity} & \thead{Tput \\ (tok/s)} & \thead{Model size \\ (GB)} \\
        \midrule
        None & 60.01 & 7.33 & 132.41 & 15.01 \\
        int4wo-64 & 58.10 & 8.25 & 268.88 & 4.76 \\
        int8wo & 59.92 & 7.34 & 216.38 & 8.04 \\
        float8wo & 59.83 & 7.37 & 213.88 & 8.03 \\
        float8dq (PerRow) & 59.86 & 7.41 & 167.13 & 8.04 \\
        float8dq (PerTensor) & 59.95 & 7.42 & 176.44 & 8.03 \\
        \bottomrule
    \end{tabular}
    \vspace{-2mm}
    \caption{\textbf{Post-training quantization (PTQ) on Llama3.1-8B}.
    Quantization reduced model size by 2-4x and increased inference
    throughput by up to 2x with minimal quantization degradation.
    All experiments use a batch size of 1 with \code{torch.compile}.}
    \label{tab:ptq}
    \vspace{-4mm}
\end{table}

\subsection{Quantization-Aware Training}
QAT refers to inserting "fake" quantization operations into the model,
which simulate the quantization process during training. This allows the model to learn to be robust
to quantization errors, improving the accuracy of the model when it is ultimately quantized post-training.
TorchAO offers simple and flexible QAT APIs that allows users to specify different quantization schemes
(e.g., INT8 dynamic activations with INT4 weights), and provides corresponding PTQ configurations to
ensure end-to-end numerical consistency (see Listing~\ref{lst:qat} in Appendix~\ref{sec:appendix_apis} for a detailed example).

TorchAO's QAT support is integrated into TorchTune's and Axolotl's fine-tuning workflows, enabling
effortless fine-tuning of a model that is intended to be quantized~(Listing~\ref{lst:qat_flow}).
During fine-tuning, all "fake" quantization operations are still performed in high precision
(e.g. BF16) even though they simulate low precision numerics (e.g. INT4). The resulting QAT checkpoint
retains the exact same model structure as the original checkpoint, and so can be used as a drop-in
replacement that offers superior post-training quantized accuracy with identical inference speeds.
TorchTune additionally provides a recipe that composes QAT with LoRA~\cite{lora} to reduce the
overheads of the extra "fake" quantization operations, yielding an 1.89x improvement in training
throughput compared to vanilla QAT~\cite{qat_lora, quantizedllama3.2}.

Recent launches of the quantized Llama 3.2 1B/3B~\cite{quantizedllama3.2} and LlamaGuard3-8B
models~\cite{llamaguard} leveraged TorchAO's QAT support to mitigate quantization degradation
in their INT4 checkpoints targeting the ARM CPU backend. This resulted in a 2-4x inference speedup,
56\% reduction in model size, and 41\% reduction in memory usage compared to the original BF16
checkpoints, while maintaining competitive performance across a wide variety of inference tasks.

\subsection{Lowering to Edge}
Models quantized using TorchAO's QAT or PTQ APIs can be easily lowered to edge backends using ExecuTorch,
which provides conversions that are adapted to natively match TorchAO's quantization patterns~(Listing~\ref{lst:qat_flow}).
This enables efficient deployment of these models on various edge backends, including Android, iOS, and CoreML.
TorchAO also provides sub-8-bit ARM CPU and Metal kernels for quantized linear and embedding
operations that can be used in eager model execution, with \code{torch.compile}, with PyTorch AOTInductor,
or in the exported model. For detailed instructions on how to export TorchAO's
quantized models to run on mobile backends, refer to the ExecuTorch README~\cite{executable_mobile_build}.

%% file: eval.tex
\section{Evaluation}

In this section, we evaluate TorchAO's FP8 training,
post-training quantization (PTQ), and quantization-aware
training (QAT) support. All experiments are performed
on 1-8 H100 GPUs, each with 96GB of HBM3 memory.

\textbf{FP8 training.} We benchmark training Llama3-8B on
the C4 dataset~\cite{c4} on 8x H100 GPUs for 100 steps
using TorchTitan. All experiments use a batch size of 1,
a sequence length of 8192, \code{torch.compile}, and per
op selective activation checkpointing (SAC). For this workload,
tensorwise scaling combined with FP8 all-gather operations
achieved a speedup of 1.25x over the BF16 baseline with on
par peak memory usage~(Table~\ref{tab:fp8}) and virtually
identical loss curves~(Appendix~\ref{sec:appendix_float8_loss_curves}).

\textbf{Post-training quantization (PTQ).} We quantize
Llama3.1-8B across a variety of PTQ settings and evaluated
the quantized models on the hellaswag and wikitext tasks
on a single H100 GPU using a batch size of 1 with
\code{torch.compile}. PTQ reduced the model size by 2-4x
and increased the inference throughput by up to 2x,
while mostly maintaining parity on hellaswag accuracy
and wikitext word perplexity across all quantization settings
compared to the baseline BF16 model~(Table~\ref{tab:ptq}).

\textbf{Quantization-aware training (QAT).} At lower
precisions such as 4-bits, quantization degradation from
PTQ alone is more pronounced and QAT becomes more effective.
We evaluate QAT by fine-tuning Llama3-8B, Llama3.1-8B,
and Llama3.2-3B on the OpenAssistant Conversations Dataset~\cite{oasst1}
on 4 H100 GPUs for 1000 steps using TorchTune. All experiments
use a batch size of 8, a learning rate of 2e-5, a weight
quantization group size of 32, and activation checkpointing.
For this workload, QAT was able to recover up to 69.8\% of
the degradation in quantized hellaswag accuracy and 82.8\%
of the degradation in quantized wikitext word perplexity~(Table~\ref{tab:qat}).


%% file: conclusion.tex
\section{Conclusion}

In this paper, we introduced TorchAO, a PyTorch-native model
optimization framework that is closely integrated into each
step of the pre-training, fine-tuning, and serving lifecycle
of LLMs. TorchAO supports popular model optimization techniques,
including FP8 training, QAT, PTQ, and 2:4 sparsity, and targets
a variety of backends including server CPU/GPU and mobile/edge.
We welcome your contributions at \url{https://github.com/pytorch/ao/}.

%% file: appendix.tex
\newpage
\appendix
\onecolumn

\section{Appendix: TorchAO FP8 Scaling Recipes}
\label{sec:appendix_fp8_scaling}
TorchAO supports the following FP8 training recipes, which have different performance/accuracy trade-offs:
\begin{itemize}
  \item \textbf{tensorwise}: This is the default recipe, with reasonable performance/accuracy trade-offs. It computes a single scaling factor for each tensor. This technique has the lowest overhead and highest performance, but is more sensitive to outliers since a single outlier anywhere in the tensor will affect the scaling of the entire tensor. This can cause higher quantization error as more values may underflow to 0. When training with FSDP, tensorwise scaling also supports an additional optimization \code{enable\_fp8\_all\_gather} which will perform the all-gathers in FSDP using FP8 to reduce communication overhead.
  \item \textbf{rowwise}: This recipe trades off a bit of performance for better accuracy. It computes scaling factors along logical rows of the left GEMM operand, and along logical columns of the right GEMM operand. Computing scaling factors for more granular slices of the tensors reduces sensitivity to outliers, improving accuracy but with a performance cost compared to tensorwise.
  \item \textbf{rowwise\_gw\_hp}: This recipe is like \code{rowwise} but it keeps the $\frac{\partial L}{\partial W}$ computation in bfloat16, as experiments have shown this to be more sensitive to lower precision, and keeping it in higher precision is better for accuracy. This recipe can also achieve higher speedups than \code{rowwise} in models where the majority of GEMMs are small than M == N == K ~= 13k.
 \end{itemize}

\section{Appendix: TorchAO APIs}
\label{sec:appendix_apis}

\begin{minipage}{\linewidth}
\begin{lstlisting}[
    style=Python,
    basicstyle=\ttfamily\small,
    frame=single,
    captionpos=b,
    keywords={convert_to_float8_training},
    label={lst:fp8_training},
    caption={\textbf{TorchAO FP8 Training Example.}}
]
torch.compile(model)
convert_to_float8_training(model)
\end{lstlisting}
\end{minipage}

\begin{minipage}{\linewidth}
\begin{lstlisting}[
    style=Python,
    basicstyle=\ttfamily\small,
    frame=single,
    captionpos=b,
    keywords={quantize_},
    label={lst:ptq},
    caption={\textbf{TorchAO PTQ Example.}}
]
# INT4 weight-only, targeting tinygemm cuda kernel
quantize_(model, Int4WeightOnlyConfig(group_size=32))

# OR INT8 dynamic activation + INT4 weight, targeting XNNPACK
quantize_(model, Int8DynamicActivationInt4Weight(group_size=32))

# OR FP8 dynamic activation + FP8 weight, targeting hopper GPUs and beyond
quantize_(model, Float8DynamicActivationFloat8WeightConfig())

# etc.
\end{lstlisting}
\end{minipage}

\begin{minipage}{\linewidth}
\begin{lstlisting}[
    style=Python,
    basicstyle=\ttfamily\small,
    frame=single,
    captionpos=b,
    keywords={quantize_, sparsify_},
    label={lst:sparsity},
    caption={\textbf{TorchAO Sparsity Example.}}
]
model = model.cuda()

# Sparse Marlin 2:4
quantize_(model, Int4WeightOnlyConfig(layout=MarlinSparseLayout()))

# OR 2:4 Sparsity
sparsify_(model, SemiSparseWeightConfig())

# OR Block Sparsity
sparsify_(model, BlockSparseWeightConfig())
\end{lstlisting}
\end{minipage}

\begin{minipage}{\linewidth}
\begin{lstlisting}[
    style=Python,
    basicstyle=\ttfamily\small,
    frame=single,
    captionpos=b,
    keywords={quantize_},
    label={lst:qat},
    caption={\textbf{TorchAO QAT Example.} TorchAO's QAT flow is separated
    into two steps, prepare and convert. During the prepare step, insert "fake"
    quantization operations into the linear and embedding modules in the model
    to simulate quantization numerics, but do not actually cast the dtypes to
    lower precision. After training, in the convert step, we replace these "fake"
    quantization operations and with real quantization operations, using the same
    code path as TorchAO's regular PTQ flow.}
]
# prepare: insert fake quantization ops
# swaps `torch.nn.Linear` with `FakeQuantizedLinear`
activation_config = FakeQuantizeConfig(torch.int8, "per_token", is_symmetric=False)
weight_config = FakeQuantizeConfig(torch.int4, group_size=32)
qat_config = IntXQuantizationAwareTrainingConfig(activation_config, weight_config),
quantize_(model, qat_config)

# train
train_loop(model)

# convert: transform fake quantization ops into actual quantized ops
# swap `FakeQuantizedLinear` back to `torch.nn.Linear` and inserts
# quantized activation and weight tensor subclasses
quantize_(model, FromIntXQuantizationAwareTrainingConfig())
quantize_(model, Int8DynamicActivationInt4WeightConfig(group_size=32))

# inference or generate (not shown)
\end{lstlisting}
\end{minipage}

\section{Appendix: Float8 Training Microbenchmarks}
\label{sec:appendix_float8_microbenchmarks}
\begin{figure}[H]
  \centering
  \advance\leftskip-0.2cm
  \includegraphics[scale=0.4]{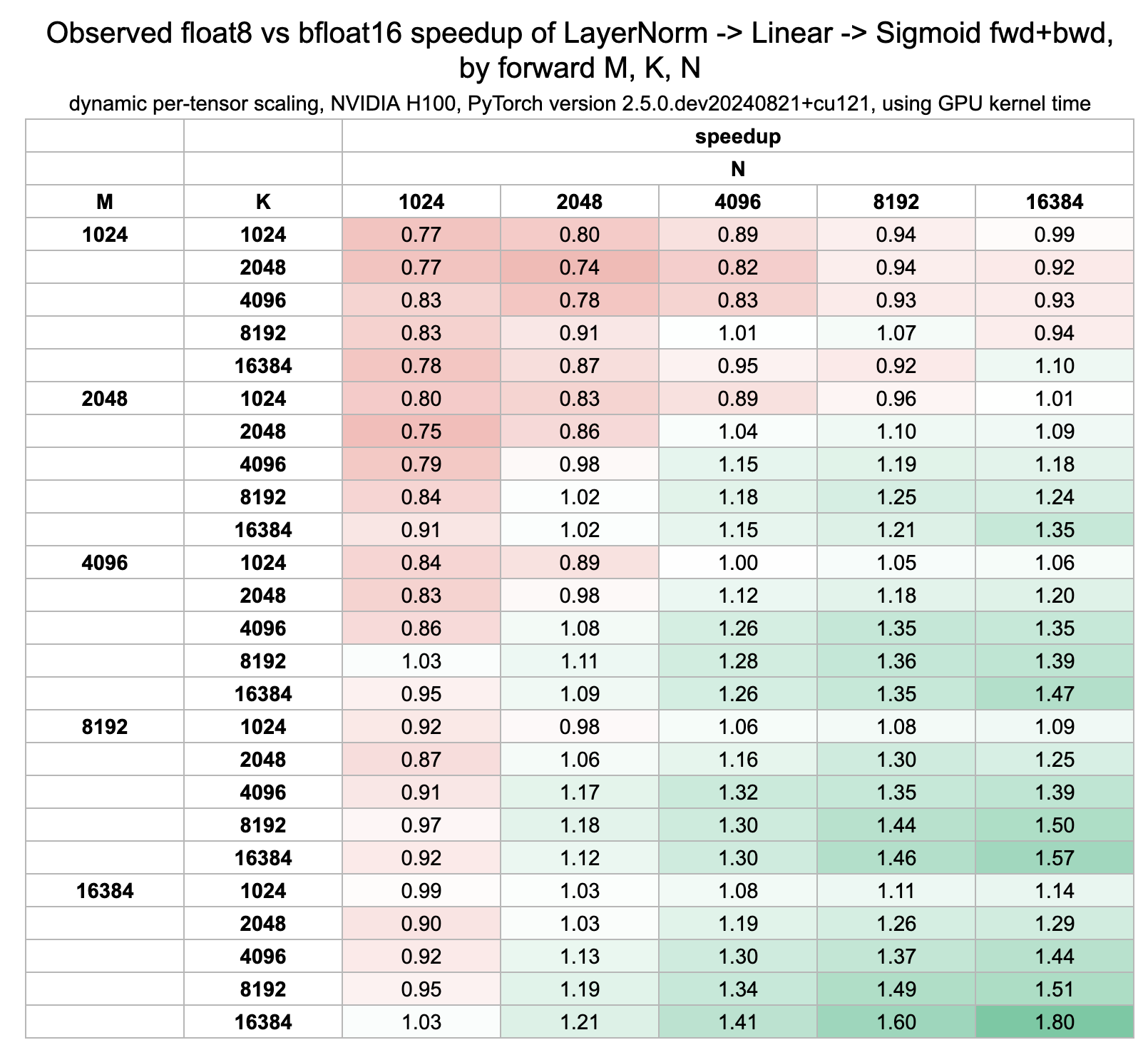}
  \caption{\textbf{Observed float8 vs bfloat16 speedup of LayerNorm + Linear + Sigmoid forward and backward pass, by forward M, N, K} A common question about FP8 training is "when is FP8 linear faster vs BF16?". Given the M, K, N of the forward pass through your linear, you can reference the table below for a microbenchmark based speedup estimate on NVIDIA H100}
  \label{image:float8_microbenchmarks}
\end{figure}

\section{Appendix: Float8 Training Loss Curves}
\label{sec:appendix_float8_loss_curves}
\begin{figure}[H]
  \centering
  \advance\leftskip-0.2cm
  \includegraphics[scale=0.4]{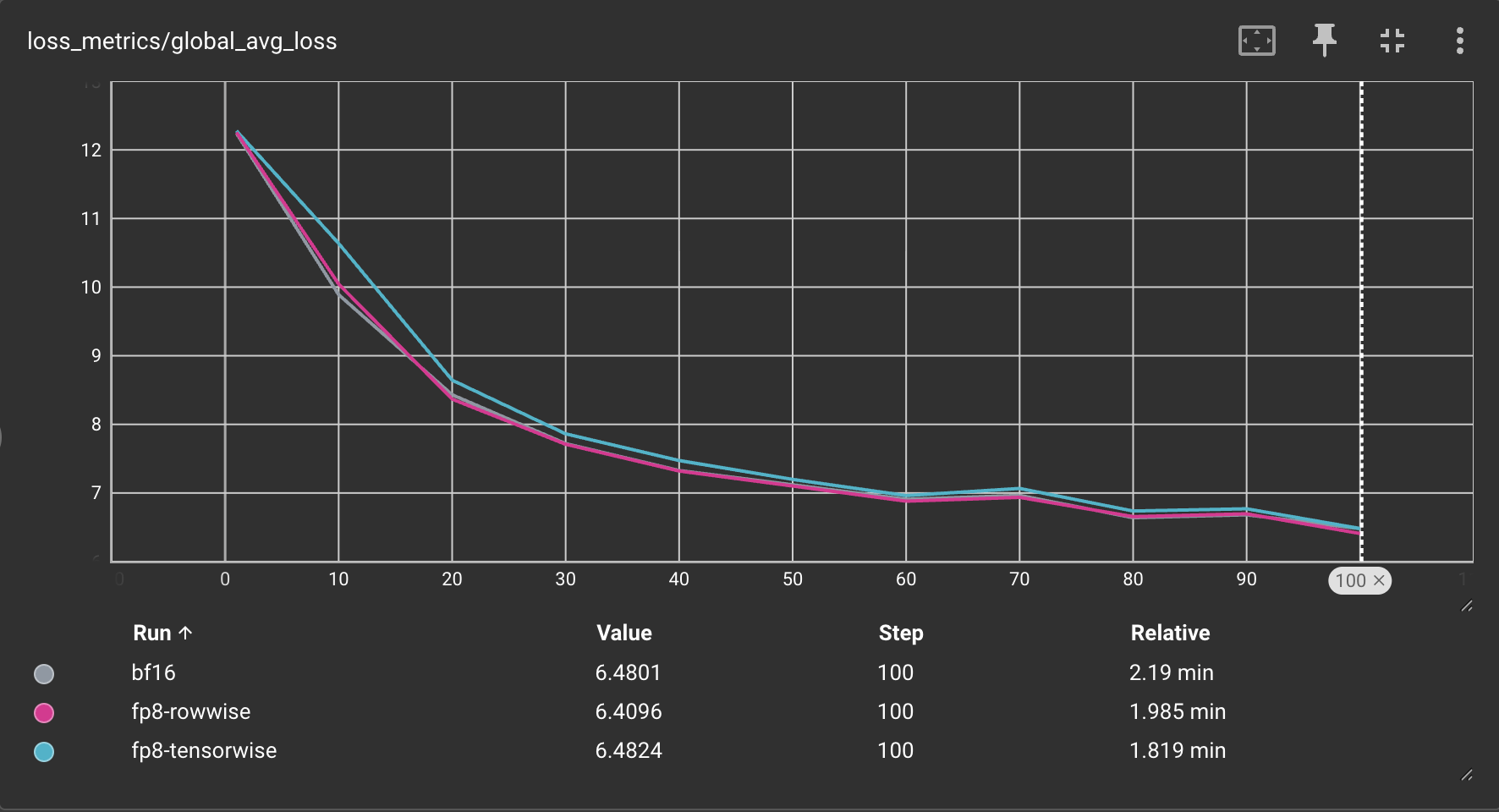}
  \caption{\textbf{Observed loss curves for float8 tensorwise and rowwise training with FSDP2 on 8xH100 GPUs, compared with the bfloat16 baseline. }}
  \label{image:float8_loss_curves}
\end{figure}

\section{Appendix: TorchAO Prototype Features}
\label{sec:appendix_prototype}
TorchAO currently has many exciting \textbf{prototype} features that are not guaranteed to be suitable for production use yet,
but are available for users to experiment with. The list of prototype features is provided below. You can find more details in the TorchAO prototypes folder \cite{torchao_prototypes}.

\begin{itemize}
  \item \textbf{AutoRound:} optimize weight rounding via signed gradient descent~\cite{autoround}
  \item \textbf{AWQ:} activation-aware weight quantization~\cite{awq}
  \item \textbf{Blockwise FP8:} FP8 blockwise quantization introduced by DeepSeek~\cite{deepseek}
  \item \textbf{float8nocompile:} accelerating eager FP8 tensorwise training with triton kernels
  \item \textbf{GaLore:} memory-efficient training with gradient low-rank projection~\cite{galore}
  \item \textbf{HQQ:} half-quadratic quantization~\cite{hqq}
  \item \textbf{MoE quantzation:} mixture of experts (MoE) quantization for inference
  \item \textbf{MX formats:} mxfp4, mxfp6, and mxfp8 for training~\cite{mx}
  \item \textbf{ParetoQ:} scaling laws in extremely low-bit LLM quantization~\cite{paretoq}
  \item \textbf{ParQ:} piecewise-affine regular quantization (QAT)~\cite{parq}
  \item \textbf{INT8 quantized training:} quantized INT8 weights or dynamic INT8 quantization
  \item \textbf{scaled\_grouped\_mm:} differentiable scaled grouped GEMM for MoE FP8 training
  \item \textbf{SmoothQuant:} remove outliers for W8A8 quantization~\cite{smoothquant}
  \item \textbf{SpinQuant:} quantization with learned rotations~\cite{spinquant}
\end{itemize}

%
%

%% file: main.bbl
\begin{thebibliography}{55}
\providecommand{\natexlab}[1]{#1}
\providecommand{\url}[1]{\texttt{#1}}
\expandafter\ifx\csname urlstyle\endcsname\relax
  \providecommand{\doi}[1]{doi: #1}\else
  \providecommand{\doi}{doi: \begingroup \urlstyle{rm}\Url}\fi

\bibitem[allenai(2025)]{c4}
allenai.
\newblock C4 dataset.
\newblock \url{https://huggingface.co/datasets/allenai/c4}, 2025.

\bibitem[Axolotl(2025)]{axolotl}
Axolotl.
\newblock Axolotl: Open source fine tuning.
\newblock \url{https://axolotl.ai/}, 2025.

\bibitem[Badri \& Shaji(2023)Badri and Shaji]{hqq}
Badri, H. and Shaji, A.
\newblock Half-quadratic quantization of large machine learning models.
\newblock \url{https://mobiusml.github.io/hqq_blog/}, November 2023.

\bibitem[Dettmers et~al.(2022)Dettmers, Lewis, Belkada, and
  Zettlemoyer]{llm_int8}
Dettmers, T., Lewis, M., Belkada, Y., and Zettlemoyer, L.
\newblock Llm.int8(): 8-bit matrix multiplication for transformers at scale.
\newblock \emph{arXiv preprint arXiv:2208.07339}, 2022.

\bibitem[Dettmers et~al.(2023)Dettmers, Pagnoni, Holtzman, and
  Zettlemoyer]{qlora}
Dettmers, T., Pagnoni, A., Holtzman, A., and Zettlemoyer, L.
\newblock Qlora: Efficient finetuning of quantized llms.
\newblock \emph{Advances in neural information processing systems},
  36:\penalty0 10088--10115, 2023.

\bibitem[executorch(2024)]{executable_mobile_build}
executorch.
\newblock Run model on android phone.
\newblock
  \url{https://github.com/pytorch/executorch/blob/main/examples/models/llama/README.md#step-4-run-benchmark-on-android-phone},
  2024.

\bibitem[executorch(2025)]{executorch}
executorch.
\newblock Executorch: End-to-end solution for enabling on-device inference
  capabilities across mobile and edge devices.
\newblock \url{https://github.com/pytorch/executorch}, 2025.

\bibitem[ExecuTorch(2025)]{torchao_executorch}
ExecuTorch.
\newblock export\_llama\_lib.
\newblock
  \url{https://github.com/pytorch/executorch/tree/main/examples/models/llama},
  2025.

\bibitem[Frantar et~al.(2025)Frantar, Castro, Chen, Hoefler, and
  Alistarh]{marlin}
Frantar, E., Castro, R.~L., Chen, J., Hoefler, T., and Alistarh, D.
\newblock Marlin: Mixed-precision auto-regressive parallel inference on large
  language models.
\newblock In \emph{Proceedings of the 30th ACM SIGPLAN Annual Symposium on
  Principles and Practice of Parallel Programming}, pp.\  239--251, 2025.

\bibitem[gemlite(2025)]{gemlite}
gemlite.
\newblock Gemlite: Triton kernels for efficient low-bit matrix multiplication.
\newblock \url{https://github.com/mobiusml/gemlite}, 2025.

\bibitem[GGML(2025)]{llama.cpp}
GGML.
\newblock Llama.cpp: Inference of meta's llama model (and others) in pure
  c/c++.
\newblock \url{https://github.com/ggml-org/llama.cpp}, 2025.

\bibitem[Google(2025)]{xnnpack}
Google.
\newblock Xnnpack.
\newblock \url{https://github.com/google/XNNPACK}, 2025.

\bibitem[Grattafiori et~al.(2024)Grattafiori, Dubey, Jauhri, Pandey, Kadian,
  Al-Dahle, Letman, Mathur, Schelten, Vaughan, et~al.]{llama3.1}
Grattafiori, A., Dubey, A., Jauhri, A., Pandey, A., Kadian, A., Al-Dahle, A.,
  Letman, A., Mathur, A., Schelten, A., Vaughan, A., et~al.
\newblock The llama 3 herd of models.
\newblock \emph{arXiv preprint arXiv:2407.21783}, 2024.

\bibitem[Guo et~al.(2025)Guo, Yang, Zhang, Song, Zhang, Xu, Zhu, Ma, Wang, Bi,
  et~al.]{deepseek}
Guo, D., Yang, D., Zhang, H., Song, J., Zhang, R., Xu, R., Zhu, Q., Ma, S.,
  Wang, P., Bi, X., et~al.
\newblock Deepseek-r1: Incentivizing reasoning capability in llms via
  reinforcement learning.
\newblock \emph{arXiv preprint arXiv:2501.12948}, 2025.

\bibitem[Han \& Han(2025)Han and Han]{unsloth}
Han, D. and Han, M.
\newblock Unsloth: Open source fine-tuning for llms.
\newblock \url{https://unsloth.ai/}, 2025.

\bibitem[Haziza et~al.(2025)Haziza, Chou, Choudhary, Wehrstedt, Massa, Yu,
  Jeong, Rao, Labatut, and Cai]{2_4_sparsity}
Haziza, D., Chou, T., Choudhary, D., Wehrstedt, L., Massa, F., Yu, J., Jeong,
  G., Rao, S., Labatut, P., and Cai, J.
\newblock Accelerating transformer inference and training with 2:4 activation
  sparsity.
\newblock \emph{arXiv preprint arXiv:2503.16672}, 2025.

\bibitem[Hu et~al.(2022)Hu, Shen, Wallis, Allen-Zhu, Li, Wang, Wang, Chen,
  et~al.]{lora}
Hu, E.~J., Shen, Y., Wallis, P., Allen-Zhu, Z., Li, Y., Wang, S., Wang, L.,
  Chen, W., et~al.
\newblock Lora: Low-rank adaptation of large language models.
\newblock \emph{ICLR}, 2022.

\bibitem[Inan et~al.(2023)Inan, Upasani, Chi, Rungta, Iyer, Mao, Tontchev, Hu,
  Fuller, Testuggine, et~al.]{llamaguard}
Inan, H., Upasani, K., Chi, J., Rungta, R., Iyer, K., Mao, Y., Tontchev, M.,
  Hu, Q., Fuller, B., Testuggine, D., et~al.
\newblock Llama guard: Llm-based input-output safeguard for human-ai
  conversations.
\newblock \emph{arXiv preprint arXiv:2312.06674}, 2023.

\bibitem[Jin et~al.(2025)Jin, Ma, Liu, Gromov, Defazio, and Xiao]{parq}
Jin, L., Ma, J., Liu, Z., Gromov, A., Defazio, A., and Xiao, L.
\newblock Parq: Piecewise-affine regularized quantization.
\newblock \emph{arXiv preprint arXiv:2503.15748}, 2025.

\bibitem[Kwon et~al.(2023)Kwon, Li, Zhuang, Sheng, Zheng, Yu, Gonzalez, Zhang,
  and Stoica]{vllm}
Kwon, W., Li, Z., Zhuang, S., Sheng, Y., Zheng, L., Yu, C.~H., Gonzalez, J.,
  Zhang, H., and Stoica, I.
\newblock Efficient memory management for large language model serving with
  pagedattention.
\newblock In \emph{Proceedings of the 29th Symposium on Operating Systems
  Principles}, pp.\  611--626, 2023.

\bibitem[Liang et~al.(2024)Liang, Liu, Wright, Constable, Gu, Huang, Zhang,
  Feng, Huang, Wang, et~al.]{torchtitan}
Liang, W., Liu, T., Wright, L., Constable, W., Gu, A., Huang, C.-C., Zhang, I.,
  Feng, W., Huang, H., Wang, J., et~al.
\newblock Torchtitan: One-stop pytorch native solution for production ready llm
  pre-training.
\newblock \emph{arXiv preprint arXiv:2410.06511}, 2024.

\bibitem[Lin et~al.(2024)Lin, Tang, Tang, Yang, Chen, Wang, Xiao, Dang, Gan,
  and Han]{awq}
Lin, J., Tang, J., Tang, H., Yang, S., Chen, W.-M., Wang, W.-C., Xiao, G.,
  Dang, X., Gan, C., and Han, S.
\newblock Awq: Activation-aware weight quantization for on-device llm
  compression and acceleration.
\newblock \emph{Proceedings of Machine Learning and Systems}, 6:\penalty0
  87--100, 2024.

\bibitem[Liu et~al.(2024)Liu, Zhao, Fedorov, Soran, Choudhary, Krishnamoorthi,
  Chandra, Tian, and Blankevoort]{spinquant}
Liu, Z., Zhao, C., Fedorov, I., Soran, B., Choudhary, D., Krishnamoorthi, R.,
  Chandra, V., Tian, Y., and Blankevoort, T.
\newblock Spinquant: Llm quantization with learned rotations.
\newblock \emph{arXiv preprint arXiv:2405.16406}, 2024.

\bibitem[Liu et~al.(2025)Liu, Zhao, Huang, Chen, Zhang, Zhao, Roy, Jin, Xiong,
  Shi, et~al.]{paretoq}
Liu, Z., Zhao, C., Huang, H., Chen, S., Zhang, J., Zhao, J., Roy, S., Jin, L.,
  Xiong, Y., Shi, Y., et~al.
\newblock Paretoq: Scaling laws in extremely low-bit llm quantization.
\newblock \emph{arXiv preprint arXiv:2502.02631}, 2025.

\bibitem[MetaAI(2025{\natexlab{a}})]{llama4}
MetaAI.
\newblock The llama 4 herd: The beginning of a new era of natively multimodal
  ai innovation.
\newblock \url{https://ai.meta.com/blog/llama-4-multimodal-intelligence/},
  2025{\natexlab{a}}.

\bibitem[MetaAI(2025{\natexlab{b}})]{quantizedllama3.2}
MetaAI.
\newblock Introducing quantized llama models with increased speed and a reduced
  memory footprint.
\newblock
  \url{https://ai.meta.com/blog/meta-llama-quantized-lightweight-models/},
  2025{\natexlab{b}}.

\bibitem[Mishra et~al.(2021)Mishra, Latorre, Pool, Stosic, Stosic, Venkatesh,
  Yu, and Micikevicius]{mishra2021accelerating}
Mishra, A., Latorre, J.~A., Pool, J., Stosic, D., Stosic, D., Venkatesh, G.,
  Yu, C., and Micikevicius, P.
\newblock Accelerating sparse deep neural networks.
\newblock \emph{arXiv preprint arXiv:2104.08378}, 2021.

\bibitem[NVIDIA(2023)]{h100}
NVIDIA.
\newblock Nvidia h100 tensor core gpu architecture.
\newblock
  \url{https://resources.nvidia.com/en-us-hopper-architecture/nvidia-h100-tensor-c},
  2023.

\bibitem[NVIDIA(2025)]{te_fp8}
NVIDIA.
\newblock Using fp8 with transformer engine.
\newblock
  \url{https://docs.nvidia.com/deeplearning/transformer-engine/user-guide/examples/fp8_primer.html#Using-FP8-with-Transformer-Engine},
  2025.

\bibitem[OpenAssistant(2025)]{oasst1}
OpenAssistant.
\newblock Openassistant conversations dataset.
\newblock \url{https://huggingface.co/datasets/OpenAssistant/oasst1}, 2025.

\bibitem[Or(2024)]{qat_lora}
Or, A.
\newblock Speeding up qat by 1.89x with lora.
\newblock
  \url{https://dev-discuss.pytorch.org/t/speeding-up-qat-by-1-89x-with-lora/2700},
  2024.

\bibitem[Or et~al.(2024)Or, Zhang, Smothers, Khandelwal, and
  Rao]{torchao_qat_blog}
Or, A., Zhang, J., Smothers, E., Khandelwal, K., and Rao, S.
\newblock Quantization-aware training for large language models with pytorch.
\newblock \url{https://pytorch.org/blog/quantization-aware-training/}, 2024.

\bibitem[PyTorch(2025)]{tensor_subclasses}
PyTorch.
\newblock Subclassing torch.tensor.
\newblock
  \url{https://docs.pytorch.org/docs/stable/notes/extending.html#subclassing-torch-tensor},
  2025.

\bibitem[PyTorch et~al.(2024)PyTorch, MobiusLabs, and SGLang]{torchao_sglang}
PyTorch, MobiusLabs, and SGLang.
\newblock Accelerating llm inference with gemlite, torchao and sglang.
\newblock \url{https://pytorch.org/blog/accelerating-llm-inference/}, 2024.

\bibitem[Qwen3(2025)]{qwen3}
Qwen3.
\newblock Qwen3: Think deeper, act faster.
\newblock \url{https://qwenlm.github.io/blog/qwen3/}, 2025.

\bibitem[Rafailov et~al.(2023)Rafailov, Sharma, Mitchell, Manning, Ermon, and
  Finn]{dpo}
Rafailov, R., Sharma, A., Mitchell, E., Manning, C.~D., Ermon, S., and Finn, C.
\newblock Direct preference optimization: Your language model is secretly a
  reward model.
\newblock \emph{Advances in Neural Information Processing Systems},
  36:\penalty0 53728--53741, 2023.

\bibitem[Rouhani et~al.(2023)Rouhani, Zhao, More, Hall, Khodamoradi, Deng,
  Choudhary, Cornea, Dellinger, Denolf, et~al.]{mx}
Rouhani, B.~D., Zhao, R., More, A., Hall, M., Khodamoradi, A., Deng, S.,
  Choudhary, D., Cornea, M., Dellinger, E., Denolf, K., et~al.
\newblock Microscaling data formats for deep learning.
\newblock \emph{arXiv preprint arXiv:2310.10537}, 2023.

\bibitem[Safaryan \& Richt{\'a}rik(2021)Safaryan and Richt{\'a}rik]{autoround}
Safaryan, M. and Richt{\'a}rik, P.
\newblock Stochastic sign descent methods: New algorithms and better theory.
\newblock In \emph{International Conference on Machine Learning}, pp.\
  9224--9234. PMLR, 2021.

\bibitem[torchao(2025{\natexlab{a}})]{torchao_kernels}
torchao.
\newblock Torchao: Low-bit arm cpu and metal kernels for linear and embedding
  ops.
\newblock \url{https://github.com/pytorch/ao/tree/main/torchao/experimental},
  2025{\natexlab{a}}.

\bibitem[torchao(2025{\natexlab{b}})]{torchao_prototypes}
torchao.
\newblock Torchao prototypes.
\newblock \url{https://github.com/pytorch/ao/tree/main/torchao/prototype},
  2025{\natexlab{b}}.

\bibitem[torchao(2025{\natexlab{c}})]{torchao_ptq_benchmarks}
torchao.
\newblock Torchao post-training quantization llama3 benchmarks.
\newblock
  \url{https://github.com/pytorch/ao/tree/main/torchao/quantization#benchmarks},
  2025{\natexlab{c}}.

\bibitem[torchao(2025{\natexlab{d}})]{torchao_sparsity_benchmarks}
torchao.
\newblock Torchao sparsity llama3 benchmarks.
\newblock
  \url{https://github.com/pytorch/ao/tree/main/torchao/sparsity#llama3},
  2025{\natexlab{d}}.

\bibitem[torchtitan(2025)]{torchtitan_async_tp}
torchtitan.
\newblock Torchtitan async tp benchmarks.
\newblock
  \url{https://github.com/pytorch/torchtitan/blob/main/benchmarks/asyncTP_llama3_h100_2025-06_torchtitan.md},
  2025.

\bibitem[torchtune(2025)]{torchtune}
torchtune.
\newblock Torchtune: Native pytorch library for llm fine-tuning.
\newblock \url{https://github.com/pytorch/torchtune}, 2025.

\bibitem[triton(2025)]{triton}
triton.
\newblock Triton: Language and compiler for writing highly efficient custom
  deep-learning primitives.
\newblock \url{https://github.com/triton-lang/triton}, 2025.

\bibitem[vLLM(2025)]{torchao_vllm}
vLLM.
\newblock Torchao documentation.
\newblock
  \url{https://docs.vllm.ai/en/latest/features/quantization/torchao.html},
  2025.

\bibitem[von Platen et~al.()von Platen, Patil, Lozhkov, Cuenca, Lambert, Rasul,
  Davaadorj, Nair, Paul, Liu, Berman, Xu, and Wolf]{hf_diffusers}
von Platen, P., Patil, S., Lozhkov, A., Cuenca, P., Lambert, N., Rasul, K.,
  Davaadorj, M., Nair, D., Paul, S., Liu, S., Berman, W., Xu, Y., and Wolf, T.
\newblock {Diffusers: State-of-the-art diffusion models}.
\newblock URL \url{https://github.com/huggingface/diffusers}.

\bibitem[Wang et~al.(2024{\natexlab{a}})Wang, He, and
  Wehrstedt]{symmetric_memory}
Wang, Y., He, H., and Wehrstedt, L.
\newblock Pytorch symmetricmemory: Harnessing nvlink programmability with ease.
\newblock
  \url{https://dev-discuss.pytorch.org/t/pytorch-symmetricmemory-harnessing-nvlink-programmability-with-ease/2798},
  2024{\natexlab{a}}.

\bibitem[Wang et~al.(2024{\natexlab{b}})Wang, He, Wright, Wehrstedt, Liu, and
  Liang]{async_tp}
Wang, Y., He, H., Wright, L., Wehrstedt, L., Liu, T., and Liang, W.
\newblock Introducing async tensor parallelism in pytorch.
\newblock
  \url{https://discuss.pytorch.org/t/distributed-w-torchtitan-introducing-async-tensor-parallelism-in-pytorch/209487},
  2024{\natexlab{b}}.

\bibitem[Wolf et~al.(2019)Wolf, Debut, Sanh, Chaumond, Delangue, Moi, Cistac,
  Rault, Louf, Funtowicz, et~al.]{hf_transformers}
Wolf, T., Debut, L., Sanh, V., Chaumond, J., Delangue, C., Moi, A., Cistac, P.,
  Rault, T., Louf, R., Funtowicz, M., et~al.
\newblock Huggingface's transformers: State-of-the-art natural language
  processing.
\newblock \emph{arXiv preprint arXiv:1910.03771}, 2019.

\bibitem[Wright et~al.(2024)Wright, Feng, Kuznetsov, and
  Guesseous]{pytorch_float8_fsdp2_blog}
Wright, L., Feng, W., Kuznetsov, V., and Guesseous, D. e.~a.
\newblock Training using float8 with fsdp2.
\newblock \url{https://pytorch.org/blog/training-using-float8-fsdp2/}, 2024.

\bibitem[Wright et~al.(2025)Wright, Shojanazeri, Kuznetsov, Vega-Myhre,
  Nadathur, Constable, Liu, Rice, Guessous, Fromm, Wehrstedt, Yu, Petersen,
  Cala, and Smith]{pytorch_float8_crusoe_blog}
Wright, L., Shojanazeri, H., Kuznetsov, V., Vega-Myhre, D., Nadathur, G.,
  Constable, W., Liu, T., Rice, T., Guessous, D., Fromm, J., Wehrstedt, L., Yu,
  J., Petersen, E., Cala, M., and Smith, C.
\newblock Accelerating large scale training and convergence with pytorch float8
  rowwise on crusoe 2k h200s.
\newblock
  \url{https://pytorch.org/blog/accelerating-large-scale-training-and-convergence-with-pytorch-float8-rowwise-on-crusoe-2k-h200s/},
  2025.

\bibitem[Xiao et~al.(2023)Xiao, Lin, Seznec, Wu, Demouth, and Han]{smoothquant}
Xiao, G., Lin, J., Seznec, M., Wu, H., Demouth, J., and Han, S.
\newblock Smoothquant: Accurate and efficient post-training quantization for
  large language models.
\newblock In \emph{International Conference on Machine Learning}, pp.\
  38087--38099. PMLR, 2023.

\bibitem[Zhao et~al.(2024)Zhao, Zhang, Chen, Wang, Anandkumar, and
  Tian]{galore}
Zhao, J., Zhang, Z., Chen, B., Wang, Z., Anandkumar, A., and Tian, Y.
\newblock Galore: Memory-efficient llm training by gradient low-rank
  projection.
\newblock \emph{arXiv preprint arXiv:2403.03507}, 2024.

\bibitem[Zheng et~al.()Zheng, Yin, Xie, Sun, Huang, Yu, Cao, Kozyrakis, Stoica,
  Gonzalez, et~al.]{sglang}
Zheng, L., Yin, L., Xie, Z., Sun, C., Huang, J., Yu, C.~H., Cao, S., Kozyrakis,
  C., Stoica, I., Gonzalez, J.~E., et~al.
\newblock Sglang: Efficient execution of structured language model programs.
\newblock \emph{https://arxiv.org/abs/2312.07104}.

\end{thebibliography}
